\documentclass{fairmeta}

\usepackage{amsmath,amssymb,amsthm}

\newtheorem{theorem}{Theorem}

\newcommand{\bx}{\boldsymbol{x}}

\newcommand{\priv}{\boldsymbol{p}}
\newcommand{\bb}{\boldsymbol{b}}

\newcommand{\softmax}{\text{softmax}}

\title{HME-PFD: Heterogeneous Mixture-of-Experts Privileged Feature Distillation}
\author[1,2,*]{Wen~Xu}
\author[1,*]{Xiaomin~Zhang}
\author[1]{Yucheng~Yang}
\author[1]{Danny~Liu}
\author[1]{Xunlong~Gui}
\author[1]{Fei~Huang}
\author[1]{Sha~Meng}
\affiliation[1]{Meta}
\affiliation[2]{University of Toronto}
\contribution[*]{Equal contribution.\quad This work was done while Wen Xu was an intern at Meta.}
\correspondence{Wen Xu at \email{realwen.xu@mail.utoronto.ca}}

\abstract{Knowledge distillation (KD) is widely used to transfer knowledge from large teacher models to smaller student models, facilitating efficient serving of machine learning systems. In the context of ads recommendation, privileged features distillation (PFD) extends KD by allowing the teacher model to leverage privileged features, i.e., features available only during training, to enhance distillation performance. However, in real-world scenarios, privileged features often have limited coverage and exhibit heterogeneous distributions, resulting in training examples of varying importance for distillation and model learning. We propose Heterogeneous Mixture-of-Experts Privileged Feature Distillation (HME-PFD), a novel approach that incorporates privileged features into the routing module of a Mixture-of-Experts-based student model. Our method introduces both hard-coded and learnable routing modules, enabling the student to dynamically determine expert assignment based on the availability and values of privileged features. This design allows the student model to better handle the heterogeneous data scenario, where privileged features are partially observed, by differentiating between high- and low-signal samples.
Extensive experiments on two post-click conversion prediction tasks demonstrate that HME-PFD consistently outperforms baseline methods. Our results show that leveraging privileged features for expert routing in the student model leads to improved model performance, even when privileged features are incomplete or sparse.}

\hypersetup{
  pdftitle={HME-PFD: Heterogeneous Mixture-of-Experts Privileged Feature Distillation},
  pdfauthor={Wen Xu, Xiaomin Zhang, Yucheng Yang, Danny Liu, Xunlong Gui, Fei Huang, Sha Meng}
}

\begin{document}
\maketitle

\section{Introduction}
Knowledge distillation (KD)~\citep{NeurIPS14_Hinton} has achieved great success in many machine learning (ML) domains such as recommendations~\citep{KDD20_Xu}, computer vision~\citep{arXiv24_Habib}, and natural language processing~\citep{arXiv24_Xu}. The high-level idea of KD is to transfer the knowledge from a large and more complex teacher model to a small student model by training the student model with additional soft labels provided by the teacher model. The original type of KD is referred to as model distillation (MD) in the literature to distinguish it from other knowledge distillation schemes, such as dataset distillation~\citep{arXiv18_Wang}.

In traditional MD, both the teacher and student models are trained using the same set of features, which are available during both training and inference. These are commonly referred to as \emph{regular features}. However, in many real-world applications, additional information, known as \emph{privileged features}, which is available during training but not at inference time, provides extra opportunities for knowledge distillation. Privileged features can provide valuable side information that, if leveraged appropriately, can significantly enhance model performance. Regular features typically include demographic information (e.g., age, gender, region), historical behaviors (e.g., click history), and contextual signals (e.g., time of day, device type). Privileged features, on the other hand, typically contain post-event information, such as post-click user behaviors in ads recommendation systems, which is not accessible during inference. For example, in ads recommendation, features such as whether a user quickly leaves the landing page and how long a user stays on the ad page after ad clicks are highly informative but unavailable at serving time. Prior work has shown that incorporating privileged features into the teacher model during training—an approach known as privileged features distillation (PFD)—can yield student models with superior serving-time performance~\citep{KDD20_Xu}. The privileged features are not directly used in the student model due to the strong correlation between the privileged features and conversion. Note that the distillation in PFD is label distillation, i.e., first training a teacher model, and then training the student model leveraging the soft labels from the trained teacher model. This is distinct from the co-training type of distillation, where the teacher and the student models are trained jointly.

\begin{figure*}[t]
\centering
\begin{subfigure}[t]{0.32\textwidth}
    \includegraphics[width=\textwidth]{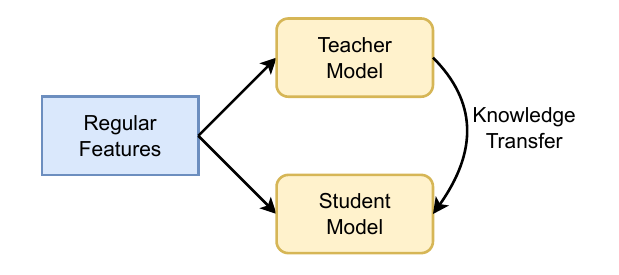}
    \caption{model distillation (MD)}
\end{subfigure}
\begin{subfigure}[t]{0.32\textwidth}
    \includegraphics[width=\textwidth]{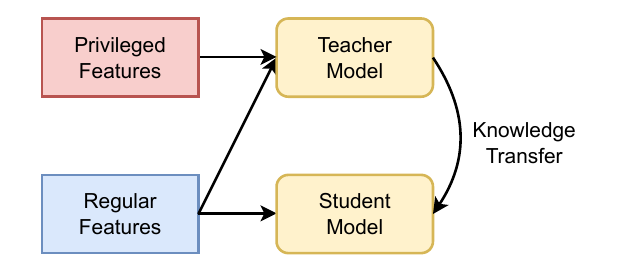}
    \caption{privileged feature distillation (PFD)}
\end{subfigure}
\begin{subfigure}[t]{0.32\textwidth}
    \includegraphics[width=\textwidth]{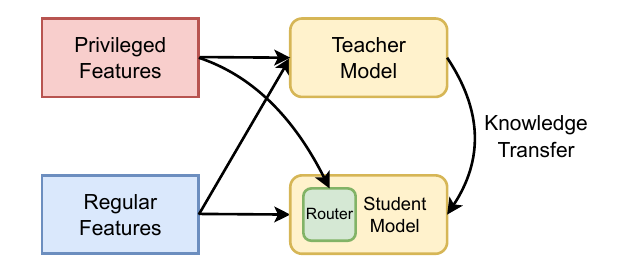}
    \caption{HME-PFD}
\end{subfigure}
\caption{Comparison of MD, PFD, and HME-PFD.}
\label{fig:illustration}
\end{figure*}


PFD assumes that all privileged features are available for all the training examples to train the teacher model. However, in realistic scenarios, the privileged features are often not available or evenly distributed for all training examples. In this heterogeneous data scenario, the absence or sparsity of privileged features leads to an uneven supervision signal during distillation: Samples with rich privileged information provide strong and informative teacher guidance, while samples without such features contribute noisier or weaker signals. Naively treating all samples equally under this heterogeneity can degrade the effectiveness of PFD, as the student model may either overfit to low-quality distilled targets or underutilize high-quality ones. Privileged features, even when partially observed, offer valuable side information to identify the reliability and informativeness of the teacher signal at the sample level. By explicitly leveraging the availability patterns and values of privileged features, the student model can differentiate between high-signal and low-signal samples, enabling more targeted learning. This observation motivates using privileged features not only to enhance the teacher but also to guide how the student processes heterogeneous training data under heterogeneous data scenarios.

These challenges arise frequently in real-world recommendation systems and can be illustrated through concrete examples. For example, due to infrastructure design or system noises, some privileged features may be absent or lost. In this scenario, the coverage of each privileged feature, i.e., the percentage of training examples with that privileged feature available, can be very low even less than 50\%. Thus, the percentage of data samples with full privileged features coverage will often be much lower. Besides, different training examples typically have different subsets of available privileged features, and, in the worst case, some training examples may not have any privileged features. Given this extra data heterogeneity characteristics in addition to the original data heterogeneity and data noise, each training example can have a very different importance for the student model training. 

To classify the traffic and allow different sub-networks for model training tailored to each traffic in the student model, we proposed \emph{Heterogeneous Mixture-of-Experts Privileged Feature Distillation} (HME-PFD). Specifically, in the student model of our \textsc{HME-PFD}, we have two tasks. The first task is the main task that predicts post-click conversion with no privileged features involved. The second task is an auxiliary knowledge distillation task leveraging both the soft labels from the teacher and the mixture of experts (MoEs) architecture with privileged features determining the routing traffic. Note that the auxiliary task is needed only in training and is discarded in serving. In serving, only the main task is deployed for prediction, and this is done without any need to access privileged features. A graphic demonstration for the comparison of how features are utilized in the teacher and student models in MD, PFD, and HME-PFD is shown in Fig.~\ref{fig:illustration}. In the demonstration, only the auxiliary knowledge distillation tasks are shown to emphasize the differences of MD, PFD, and HME-PFD in distillation.

In our \textsc{HME-PFD}, the MoEs sub-networks can be of any model architecture, for which we choose the standard multi-layer perceptrons (MLPs). The routing module of the MoEs determines the traffic by the privileged features of the training examples, either via hard-coded rules or learnable modules using those privileged features. As we mentioned earlier, PFD does not directly utilize any privileged features in the student model and generally assumes uniform availability and quality of privileged features. Therefore, PFD has no need to reason about routing or traffic partitioning at all. Once privileged features are sparse, heterogeneous, and sample-dependent, how routing is done directly controls whether the student model amplifies high-quality supervision or dilutes it with noisy signals. Our work is the first to (i) identify routing as the correct abstraction for handling heterogeneous privileged-feature coverage in student-side distillation, and (ii) systematically study routing design choices, from deterministic availability-based rules to expressive learnable gates, within a unified framework. The contribution is not any single routing architecture, but the demonstration that routing flexibility is one way to adapt distillation strength to supervision heterogeneity.

The full contributions of this paper are summarized as follows:
\begin{itemize}
    \item We formally identified and analyzed the problem of heterogeneous and sparse privileged features in the context of knowledge distillation.
    \item We propose \emph{Heterogeneous Mixture-of-Experts Privileged Feature Distillation} (\textsc{HME-PFD}), a new distillation framework that explicitly addresses the \emph{heterogeneous data} setting in which privileged features are heterogeneous and partially observed. Our approach equips the student model to differentiate training samples with heterogeneous supervision quality.
    \item We develop a flexible routing design for \textsc{HME-PFD} that accommodates heterogeneous privileged-feature characteristics, including partial availability, varying scales, and complex interactions. Our framework supports both deterministic, hard-coded routing rules, suitable when privileged features act as clear availability indicators, and a family of learnable routing modules, such as linear gating, multi-layer perceptrons, residual networks, and attention-based architectures, which enable adaptive traffic partitioning when simple rules are insufficient.
    \item We conduct extensive experiments on large-scale industrial ads recommendation tasks under multiple deployment scenarios. The results demonstrate that \textsc{HME-PFD} consistently outperforms strong baselines and provide systematic insights into the effects of key design choices, including the number of experts, the number of selected experts, the expert network architectures, the routing paradigm, etc.
\end{itemize}

\section{Related Work}\label{sec:related_work}

Utilizing distilled knowledge from a much larger model, knowledge distillation (KD) was originally proposed to train compact models that are easier to deploy while retaining competitive performance~\citep{NeurIPS14_Hinton}. The core idea of KD is to transfer knowledge from a large and complex \emph{teacher} model to a smaller \emph{student} model by training the student to match the teacher’s soft output probabilities, in addition to minimizing its empirical loss with respect to the ground-truth labels. Since the seminal work of~\citep{NeurIPS14_Hinton}, a rich body of literature has emerged to improve KD from different perspectives, including incorporating structural knowledge via contrastive representation distillation~\citep{ICLR20_Tian}, decoupling target and non-target class information~\citep{CVPR22_Zhao}, designing curriculum-based temperature schedules~\citep{AAAI23_Li}, and formulating distillation objectives using $\alpha$--$\beta$ divergences, such as ABKD~\citep{ICML25_Wang}. Other distillation paradigms have also been explored, notably dataset distillation~\citep{arXiv18_Wang}. Following common terminology in the literature, we refer to the original form of KD, where both teacher and student are trained on the same data representation, as \emph{model distillation} (MD), to distinguish it from these alternative paradigms.

In the aforementioned distillation methods, both the teacher and the student rely on the same set of input features, which are available at both training and inference time. These features, commonly referred to as \emph{regular features}, are standard in many machine learning settings where the feature space remains consistent between training and serving. In contrast, additional information may be available only during training but not during inference. Learning frameworks that exploit such information fall under the paradigm of \emph{learning using privileged information} (LUPI)~\citep{NN09_Vapnik, JMLR15_Vapnik}. In LUPI, the teacher model is granted access to privileged features at training time, with the goal of transferring this additional knowledge to a student model that operates without them at inference.

In the domain of ads recommendation, both regular and privileged features are commonly available during training. Privileged feature distillation (PFD) was introduced in~\citep{KDD20_Xu}, where a teacher model is trained using both regular and privileged features, and its predictions are subsequently distilled into a student model that only consumes regular features at serving time. A key distinction between PFD and MD is that, in MD, both teacher and student share the same feature space, whereas in PFD the teacher has exclusive access to privileged features. Theoretical analysis on linear models, together with extensive empirical studies, was provided in~\citep{NeurIPS22_Yang}, which characterized when and why PFD improves student performance, revealed a non-monotonic effect of privileged feature strength, and explained these behaviors through variance reduction and variance inflation mechanisms.

However, existing PFD approaches typically assume that privileged features are fully observed across training data. In practice, privileged features may have limited or uneven coverage in real-world large-scale recommendation systems, leading to heterogeneous supervision quality across training examples. Motivated by this observation, we extend PFD by incorporating mixture-of-experts (MoEs)~\citep{NC91_Jacobs, ICLR17_Shazeer} as the core architecture of the student model, and by explicitly leveraging privileged features within the MoE routing module to better handle data heterogeneity and traffic classification under partial privileged-feature availability.

\begin{figure*}[t]
\centering
\includegraphics[width=0.85\linewidth]{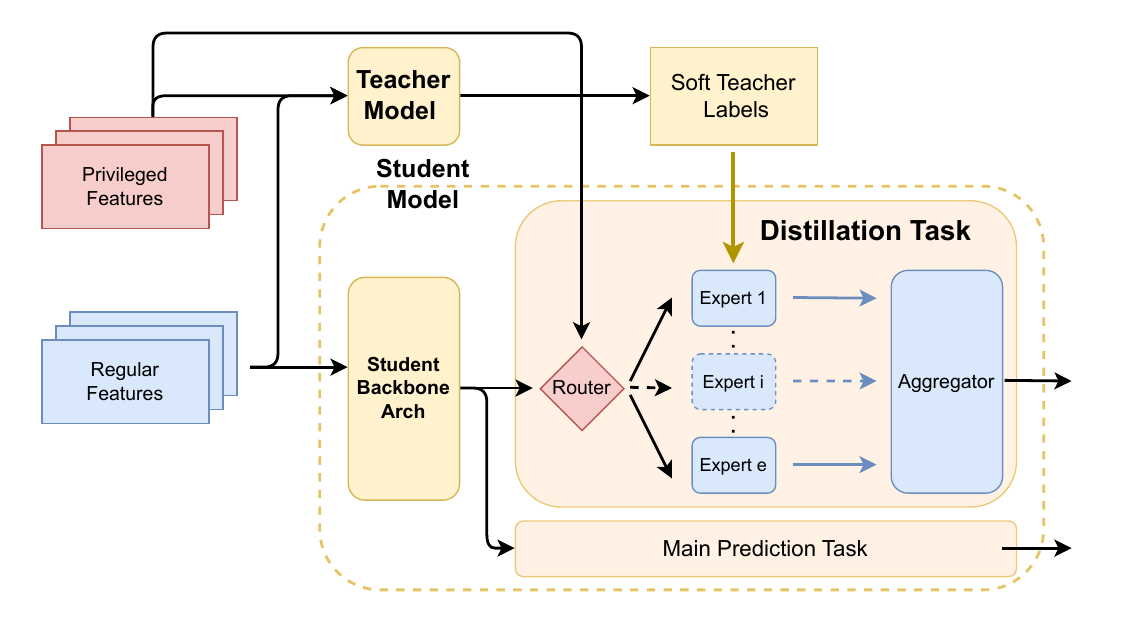}
\caption{A demonstration of HME-PFD applied to the student model.}
\label{fig:HME-PFD}
\end{figure*}

\section{Heterogeneous Mixture-of-Experts Privileged Feature Distillation}
\label{sec:HME-PFD}

In this section, we formalize the learning problem and describe the system design and model architecture of \textsc{HME-PFD}. Our goal is to train a student model for a primary prediction task (e.g., conversion prediction) using knowledge distilled from a teacher model that has access to privileged features. 

\subsection{Learning Problem Formulation}

We begin by presenting the mathematical formulation of the learning problem underlying \textsc{HME-PFD}. Each training example is represented by a tuple $(\bx, \bx^\star, y)$, where $\bx \in \mathcal{X} \subseteq \mathbb{R}^{n_1}$ denotes the \emph{regular features}, $\bx^\star \in \mathcal{X}^\star \subseteq \mathbb{R}^{n_2}$ denotes the \emph{privileged features}, and $y \in \{0,1\}$ is the binary ground-truth label. The training dataset is given by $\{(\bx_i, \bx_i^\star, y_i)\}_{i=1}^{N}$.

Let $\ell: \mathbb{R} \times \mathbb{R} \rightarrow \mathbb{R}_{+}$ denote a point-wise loss function that measures the discrepancy between a prediction $\hat{y}$ and the corresponding label $y$, i.e., $\ell(y,\hat{y})$. The specific choice of $\ell$ is task-dependent; for binary classification, common choices include cross-entropy loss or normalized cross-entropy loss. We denote the parameters of the teacher and student models by $W_t$ and $W_s$, respectively.

The teacher model is trained by minimizing its empirical risk and yields a predictive function $f(\cdot; W_t)$, which produces soft predictions $f(\bx, \bx^\star; W_t)$ using both regular and privileged features during training. Given these soft labels, the student model is trained by solving the following optimization problem:
\begin{align}
\min_{W_s} & \quad \frac{1}{N}\sum_{i=1}^{N} \Big[ (1-\lambda)\,
\ell_{\text{main}}\big(y_i, f(\bx_i; W_s)\big) + \lambda\, \ell_{\text{distill}}\big(f(\bx_i, \bx_i^\star; W_t), f(\bx_i, \bx_i^\star; W_s)\big) \Big],
\end{align}
where $\ell_{\text{main}}(\cdot)$ denotes the loss of the main prediction task, and $\ell_{\text{distill}}(\cdot)$ denotes the distillation loss of the auxiliary knowledge distillation task that encourages the student to match the teacher’s soft outputs. The hyperparameter $\lambda \in [0,1]$ balances the main task and the auxiliary distillation objective.

The core challenge is that the privileged features $\bx^\star$ are often sparse and heterogeneously distributed. For many examples, $\bx^\star$ may be partially or entirely missing. Our objective is to learn a student model that can effectively predict $y$ using only the regular features $\bx$ at inference time, while leveraging the \emph{heterogeneous} privileged information $\bx^\star$ during training. While feature imputation techniques may be employed to handle missing privileged features~\citep{AIR20_Lin}, advanced methods such as GAN-based~\citep{ICML18_Yoon} or diffusion-based imputation~\citep{ICLR25_Zhang} introduce significant computational overhead, especially when the number of privileged features is large.

We provided theoretical intuitions for PFD and HME-PFD in Appendix~\ref{appendix:scenario-distillation}. There we study the auxiliary distillation objective with a fixed teacher and a fixed student representation. We compare the minimum population squared loss of a single auxiliary head with that of experts routed by privileged-feature scenario. Routing gives a strict improvement exactly when knowing the scenario changes the conditional mean of the teacher target for a set of examples of positive probability. We evaluate the deployed main head separately on ground-truth labels.

\subsection{System Design}

To address the issue of privileged feature heterogeneity, we propose the \textsc{HME-PFD} framework, which is built upon a Mixture-of-Experts (MoE) architecture. \textsc{HME-PFD} follows a standard privileged features distillation paradigm, where teacher model consumes both regular and privileged features, while student model only takes regular features as direct inputs during training. Both teacher and student model share a similar deep, multi-layer backbone architecture detailed in~\citep{zhang2022dhendeephierarchicalensemble}. We implement \textsc{HME-PFD} as part of the student model architecture, which is built upon common design principles used in large-scale recommendation systems~\citep{arXiv19_Naumov, FairScale2021}.

The student model works as follows: first, the student's feature-extraction backbone architecture processes all regular features, including user features, item features, interaction features, and other relevant signals as illustrated in Fig.~\ref{fig:HME-PFD}. For each input instance, this backbone architecture produces a fixed-dimensional embedding, where the dimension is application-dependent, with typical choices drawn from $\{64,128,256,512\}$ depending on model capacity, dataset scale, and task complexity. 

Following the multi-task multi-label learning design, we construct multiple prediction heads consuming the shared embedding from the backbone architecture, including main task architecture and other auxiliary tasks, to support different learning objectives. The main task head typically predicts either the click-through rate (CTR), defined as the probability of a user clicking an item upon impression, or the post-click conversion rate (CVR), defined as the probability of a desired action following a click~\citep{fb_ctr}. This main task is supervised using ground-truth labels from click-through or conversion during offline training, while the distillation task is supervised by soft teacher labels.

Different from single MLP blocks used in PFD, in \textsc{HME-PFD}, we introduce an auxiliary \emph{distillation task} implemented via a mixture-of-experts (MoE) architecture~\citep{NC91_Jacobs, ICLR17_Shazeer}. Each MoE module consists of a routing function, a set of expert networks, and a loss aggregation mechanism. Motivated by the heterogeneity induced by incomplete privileged features, we explicitly leverage privileged features within the routing module to guide expert selection. In terms of routing module design, we consider both hard-coded routing strategies and learnable routing functions.

Depending on the routing decisions from the router, each input embedding is dispatched to one or multiple experts. The expert networks can adopt arbitrary architectures; following common practice, we instantiate them as multi-layer perceptrons (MLPs), though the framework readily generalizes to other neural architectures. Finally, the losses produced by individual experts are combined by a loss aggregation module, typically via a weighted sum or weighted average, yielding a single scalar loss for the auxiliary distillation task. This design ensures that \textsc{HME-PFD} remains a single auxiliary objective, avoiding the need for additional task-level loss weighting.

Importantly, \textsc{HME-PFD} differs fundamentally from prior privileged feature distillation (PFD) approaches~\citep{KDD20_Xu}. While PFD restricts the use of privileged features to the teacher model, our method additionally exploits privileged features within the student model through the MoE routing mechanism during training, dealing with data heterogeneity issues. A schematic illustration of the student architecture in \textsc{HME-PFD} is provided in Fig.~\ref{fig:HME-PFD}.

\section{Routing Module Design}
\label{sec:router_design}

In this section, we describe the design of the routing module used in the mixture-of-experts (MoE) component of \textsc{HME-PFD}. The effectiveness of \textsc{HME-PFD} hinges on the design of the routing module, which is responsible for assigning training instances to the appropriate experts based on the privileged features. The router's goal is to differentiate examples with distinctive characteristics, enabling the MoE to learn specialized representations.

Let $\bx_{\text{embed}} \in \mathbb{R}^{n}$ denote the embedding of a raw input instance, where $n$ is the embedding dimension. In ads recommendation systems, this embedding is produced by a shared feature-extraction backbone that is jointly used by the main prediction task and multiple auxiliary tasks to facilitate representation learning. Our MoE-based distillation component in \textsc{HME-PFD} can be viewed as one such auxiliary task. Specifically, the MoE takes $\bx_{\text{embed}}$ as input and produces a distillation loss by comparing the student’s predictions with the teacher’s soft labels.

As discussed in Section~\ref{sec:HME-PFD}, \textsc{HME-PFD} leverages privileged features to guide routing decisions in the MoE. We denote the privileged features used for routing by $\priv \in \mathbb{R}^{p}$, where $p$ is the number of selected privileged features. Note that $\priv$ in general can be constructed by any subset of the full privileged features from vector $\bx^\star \in \mathbb{R}^{n_2}$. The routing module dispatches each input to one or more experts, where each expert is instantiated as a neural network. The total number of experts is denoted by $e$.

The routing workflow proceeds as follows. For each training instance, the routing module consumes the corresponding privileged features $\priv$ and produces a score for each expert. Based on these scores, the input embedding $\bx_{\text{embed}}$ is routed to the expert with the highest score (top-$1$ routing) or, more generally, to the top-$k$ experts. Formally, the routing module defines a function
\[
g: \mathbb{R}^{p} \rightarrow \mathbb{R}^{e},
\]
which maps privileged features to expert scores. 

In the following, we introduce two main categories of routing strategies: (i) a hard-coded routing strategy, and (ii) a variety of learnable routing modules, including linear layers, multi-layer perceptrons, and attention-based architectures. The routing choice depends on the availability of prior knowledge about how privileged features should guide traffic allocation.
When such prior knowledge exists, the routing function $g(\cdot)$ can be explicitly defined to leverage this information. Otherwise, $g(\cdot)$ can be parameterized as a learnable function and optimized jointly with the student model.





\subsection{Hard-Coded Routing}
\label{subsec:hard-coded}


The simplest routing strategy is to define the routing function deterministically based on the availability of privileged features. Consider using a single privileged feature $\priv = f_1 \in \mathbb{R}$ to route inputs in an MoE with two experts, a hard-coded routing rule can be defined as follows: If the privileged feature $f_1$ is missing for a data point, the input is routed to the first expert; Otherwise, it is routed to the second expert. This rule can be formalized as:
\begin{align}
g_{\text{hard}}(\priv) =
\begin{cases}
[1,\,0]^\top, & \text{if } \priv = \text{NULL}, \\
[0,\,1]^\top, & \text{otherwise},
\end{cases}
\label{eqn:router:hard-coded}
\end{align}
where the routing score vector determines expert selection by choosing the expert with the larger score. For instance, if the coverage of the privileged feature $f_1$ is approximately $20\%$ in the training dataset, then about $80\%$ of the data points will be routed to the first expert, while the remaining $20\%$ will be routed to the second expert. This hard-coded routing strategy enables a simple yet effective separation of samples based on privileged feature availability, which might improve the student model performance if indeed the existence of $f_{1}$ reflects the signal strength of each training example.

The routing rule in~\eqref{eqn:router:hard-coded} can be readily extended to settings with multiple privileged features or more complex deterministic rules. However, in practice, the number of experts may exceed two, and the optimal routing logic may not be obvious or easily expressible by hand-crafted rules. In other words, hard-coded routing may not scale, even based on the existence of the privileged features. For example, if $p$ privileged features are utilized in routing, the total number of possible existence patterns is $2^{p}$, which in turn requires $2^{p}$ experts if we assign one expert for each pattern. This exponential dependence on the number of privileged features introduces significant inefficiency in model architecture design, and is prohibitive when $p$ becomes large. Furthermore, this type of hard-coded routing cannot capture the confidence level or any complex interactions among the features.

These limitations motivate the use of learnable routing modules, which we describe in detail in Section~\ref{sec:learnable}. Allowing learnable routing modules, such as a simple linear layer or complex neural networks, can automatically determine scores for each expert and figure out which expert any data point should be routed to.

\subsection{Learnable Routing Modules}\label{sec:learnable}

When no clear deterministic rule is available for routing based on privileged features, the routing module can instead be parameterized as a learnable function. Learnable routing modules offer greater flexibility and can adaptively discover traffic patterns induced by heterogeneous privileged features.

\subsubsection{Linear Layer}
A simple and widely used choice for learnable routing is a linear gating function~\citep{ICLR17_Shazeer, NeurIPS22_Zhou}, where each expert is associated with a learnable weight vector over the routing privileged features, and the routing score is computed as a weighted sum followed by a bias term. Formally, the routing function is defined as
\begin{align}
g_{\text{linear}}(\priv) = \softmax(W_{\text{linear}}\priv + \bb),
\label{eqn:linear}
\end{align}
where $W_{\text{linear}} \in \mathbb{R}^{e \times p}$ is the weight matrix, $\bb \in \mathbb{R}^{e}$ is the bias vector, and $\softmax(\cdot)$ denotes the standard softmax operator. The learnable parameters of this routing module are $W_{\text{linear}}$ and $\bb$.

The output $g_{\text{linear}}(\priv) \in \mathbb{R}^{e}$ assigns a routing score to each expert, and the input embedding $\bx_{\text{embed}}$ is dispatched to the expert with the highest score (or the top-$k$ experts).

\subsubsection{Multi-Head Attention}

Multi-head attention~\citep{NeurIPS17_Vaswani} provides a powerful mechanism for capturing interactions across multiple representation subspaces. We adapt self-attention for routing by treating each privileged feature as a one-dimensional token.

Let $\{x_1, \dots, x_m\}$ denote the tokenized privileged features. For each token $x_i$, we compute
\begin{align}
q_i = W_q x_i, \quad
k_i = W_k x_i, \quad
v_i = W_v x_i,
\end{align}
where $W_q$, $W_k$, and $W_v$ are learnable projection matrices. The attention weights and context vectors are computed as
\begin{align}
\alpha_j &= \softmax(K^\top q_j), \\
c_j &= V \alpha_j,
\end{align}
where $K = [k_1,\dots,k_m]$ and $V = [v_1,\dots,v_m]$. The outputs of multiple heads are concatenated and processed by a final linear layer. The routing function is defined as
\begin{align}
g_{\text{attention}}(\priv) =
\softmax(W_{\text{linear}} f_{\text{attention}}(\priv) + \bb),
\label{eqn:attention}
\end{align}
where $f_{\text{attention}}(\priv)$ denotes the multi-head attention output and $W_{\text{linear}} \in \mathbb{R}^{e \times h}$ projects it to expert scores.

\subsubsection{Other Learnable Layers} Other nonlinear learnable routing modules are also possible. In particular, multi-layer perceptrons (MLPs) can capture nonlinear feature dependencies, and residual networks (ResNet) may enable increasingly expressive nonlinear transformations with improved optimization stability. These architectures differ primarily in their capacity and inductive biases, but share the common goal of learning a flexible mapping from privileged features to expert scores. In our framework, all such nonlinear routing modules are interchangeable and are trained jointly with the student model using the same objective. As shown in our experiments, while more expressive routing functions can yield modest performance gains in certain settings, simpler designs such as linear or shallow nonlinear routing often capture most of the benefit, highlighting that the effectiveness of \textsc{HME-PFD} stems from privileged-feature–aware routing itself rather than any specific routing architecture.

\subsection{Extensions}
In general, the routing module may be instantiated by any function
$g: \mathbb{R}^{p} \rightarrow \mathbb{R}^{e}$. Beyond the routing architectures considered above, several extensions may further improve performance, including hierarchical gating~\citep{ICLR17_Shazeer}, gating dropout~\citep{ICML22_Liu}, and stochastic routing noise for enhanced exploration~\citep{ICLR17_Shazeer}. Since the optimal routing strategy is application-specific, we leave the choice of routing architecture to domain knowledge and empirical validation.

In all proposed routing modules, expert selection is performed via top-$1$ or top-$k$ routing, corresponding to sparsely gated MoEs~\citep{ICLR17_Shazeer}. In Appendix~\ref{appendix:paradigms}, we further explore alternative paradigms that do not rely on top-$k$ selection, including soft MoEs~\citep{ICLR24_Puigcerver}, where inputs contribute to all experts with learned weights, and expert-choice routing~\citep{NeurIPS22_Zhou}. Detailed results are deferred to the Appendix.

\section{Experiments}
\label{sec:experiments}

We conduct a series of extensive experiments to evaluate the effectiveness of \textsc{HME-PFD} framework. Our evaluation is designed to address the following four key aspects:
\begin{itemize}
    \item \textbf{Overall Effectiveness}: We compare \textsc{HME-PFD} against \textsc{PFD}, and non-distilled baselines to quantify performance improvements on large-scale industrial datasets.
    \item \textbf{Routing Mechanism Design}: We evaluate the impact of different routing strategies, including deterministic hard-coded rules and learnable architectures (linear gating, multi-layer perceptrons, and multi-head attention).
    \item \textbf{Architecture Configuration}: We conduct ablation studies on the number of experts and expert network capacity to understand their effects on model performance.
    \item \textbf{Expert Selection Strategy}: We analyze how the number of selected experts (top-k routing) affects prediction accuracy.

\end{itemize}

\subsection{Experimental Setup}
We use two large-scale industrial datasets for post-click conversion rate (CVR) prediction tasks, which correspond to the final-stage, fine-grained ranking models in a cascaded ads recommendation system. These datasets consist of billions of user interactions and are representative of two real-world advertising scenarios, named \textsc{CVR\_Large} and \textsc{CVR\_Small}, each targeting a different conversion prediction task. CVR is defined as the probability that a user completes a desired action, e.g., purchase or installation, after clicking an ad. The privileged features include post-click user engagement signals, which are only available during training. 

\paragraph{Evaluation Metric.}
We adopt \emph{Normalized Entropy} (NE), also known as normalized cross-entropy loss, which is a standard offline evaluation metric for recommendation and ranking systems~\citep{ADKDD14_He, KDD22_Zhang}. NE normalizes the average binary cross-entropy loss by that of a constant baseline predictor, enabling meaningful comparison across datasets with different label distributions. Formally,
\begin{align}
\mathrm{NE}
=
\frac{
\frac{1}{N}\sum_{i=1}^{N}
\big( y_i \log p_i + (1-y_i)\log(1-p_i) \big)
}{
p_{\text{data}} \log(p_{\text{data}}) + (1-p_{\text{data}})\log(1-p_{\text{data}})
},
\end{align}
where $N$ is the total number of samples, $y_i \in \{0,1\}$ is the ground-truth label, $p_i$ is the predicted probability for the positive class, and $p_{\text{data}}$ denotes the empirical positive rate in the dataset.

The normalization is crucial for robust comparison: when $p_{\text{data}}$ is close to $0$ or $1$, a constant prediction of $p_{\text{data}}$ achieves low cross-entropy loss. NE mitigates this issue by accounting for the class imbalance. Clearly, lower NE values indicate better model performance.

\paragraph{Baselines.}
We consider the \textsc{Plain} baseline where a student model is trained without any knowledge distillation, using only ground-truth supervision. We consider the following routing module design in our \textsc{HME-PFD}.
\begin{itemize}
    \item \textsc{Hard-coded}: A simple deterministic routing rule based on privileged feature availability. Specifically, when $e=2$, inputs without privileged features are routed to the first expert, while others are routed to the second expert.
    \item \textsc{Linear}: Learnable routing using a linear gating function as defined in~\eqref{eqn:linear}.
    \item \textsc{MLP}: Routing via a multi-layer perceptron.
    \item \textsc{ResNet}: Routing via a residual neural network.
    \item \textsc{Attention}: Routing via multi-head self-attention as defined in~\eqref{eqn:attention}.
\end{itemize}
Unless stated otherwise, we use $e=2$ experts and two privileged dwell-time features for routing.

\begin{table}[t]
  \caption{NE difference results for \textsc{CVR\_Large} and \textsc{CVR\_Small} models with two experts (Lower NE indicates better performance)}
  \label{table:NE}
  \centering
  \begin{tabular}{lcc}
    \toprule
    \multicolumn{1}{c}{} 
    & \textsc{CVR\_Large} 
    & \textsc{CVR\_Small} \\
    \cmidrule(r){2-2}
    \cmidrule(r){3-3}
    Routing module & NE diff & NE diff \\
    \midrule
    \textsc{Plain}        & $0.00\%$  & $0.00\%$   \\
    \textsc{Hard-coded}  & $-0.07\%$ & $-0.31\%$  \\
    \textsc{Linear}      & $-0.09\%$ & $-0.27\%$  \\
    \textsc{MLP}         & $-0.06\%$ & $-0.42\%$  \\
    \textsc{ResNet}      & $-0.09\%$ & $-0.32\%$  \\
    \textsc{Attention}  & $-0.07\%$ & $-0.25\%$  \\
    \bottomrule
  \end{tabular}
\end{table}

\begin{table}[t]
  \caption{NE  difference results for \textsc{CVR\_Large} with varying numbers of experts}
  \label{table:varying_experts:Web}
  \centering
  \begin{tabular}{lccc}
    \toprule
    \multicolumn{1}{c}{}   
    & $2$ experts    
    & $3$ experts 
    & $4$ experts \\
    \cmidrule(r){2-2}
    \cmidrule(r){3-3}
    \cmidrule(r){4-4}
    Routing module & NE diff & NE diff & NE diff \\
    \midrule
    \textsc{Plain}  & $0.00\%$ & N/A       & N/A       \\
    \textsc{Hard-coded}  & $-0.07\%$ & N/A       & N/A       \\
    \textsc{Linear}     & $-0.09\%$ & $-0.08\%$ & $0.11\%$  \\
    \textsc{MLP}        & $-0.06\%$ & $-0.07\%$ & $0.10\%$  \\
    \textsc{ResNet}     & $-0.09\%$ & $-0.06\%$ & $0.01\%$  \\
    \textsc{Attention} & $-0.07\%$ & $-0.03\%$ & $0.49\%$  \\
    \bottomrule
  \end{tabular}
\end{table}

\subsection{Results on \textsc{CVR\_Large}}

For the \textsc{CVR\_Large} dataset, each expert is implemented as an MLP with layer dimensions $(512,1)$. Models are trained on $14$ consecutive days of data and evaluated on the following day. Table~\ref{table:NE} reports the relative difference of the test NE for different routing strategies.

We observe that both hard-coded routing and learnable routing consistently outperform the \textsc{Plain} baseline, demonstrating the benefit of leveraging privileged features for traffic differentiation. Among the learnable routing methods, \textsc{Linear} and \textsc{ResNet} achieve the strongest performance under this setting. Given its simplicity and competitive accuracy, linear routing provides an attractive trade-off between performance and computational overhead. Note that although the relative improvements range from $0.06\%$ to $0.09\%$ in NE, such gains are practically meaningful at an industrial scale.

\subsection{Results on \textsc{CVR\_Small}}

We trained the models on $30$ days of data and evaluated on the subsequent day for the \textsc{CVR\_Small} dataset. Results of the relative difference of the test NE are summarized in Table~\ref{table:NE}. 

Similar to the \textsc{CVR\_Large} setting, all routing-based approaches outperform the \textsc{Plain} baseline. Under this configuration, \textsc{MLP}-based routing achieves the best performance, suggesting that more expressive routing functions are beneficial when user behavior signals exhibit higher complexity. If computational resources are constrained, linear routing remains a reasonable alternative with a modest performance trade-off. The observed NE improvements range from $0.25\%$ to $0.42\%$ in NE, which can translate to substantial gains in large-scale deployment.

\subsection{Effect of the Number of Experts}

We study the impact of the number of experts by varying $e \in \{2,3,4\}$. Results for the \textsc{CVR\_Large} dataset are reported in Table~\ref{table:varying_experts:Web}. Increasing the number of experts from $2$ to $3$ yields comparable performance, while further increasing to $4$ leads to noticeable degradation. The poor behavior when $e=4$ may stem from under-utilization of all the experts or the increased optimization difficulty due to a larger number of experts. This phenomenon indicates that a small number of experts is sufficient in this setting.

\begin{table}[t]
  \caption{NE difference results for \textsc{CVR\_Large} with varying expert architectures}
  \label{table:varying_expert_architectures:Web}
  \centering
  \begin{tabular}{ll|ll}
    \toprule
    Layer dimensions & NE diff 
    & Layer dimensions & NE diff \\
    \midrule
    $[256, 1]$        & $0.00\%$
    & $[256, 256, 1]$   & $0.07\%$ \\
    $[512, 1]$        & $-0.02\%$
    & $[512, 512, 1]$   & $0.04\%$ \\
    $[1024, 1]$       & $0.12\%$
    & $[1024, 1024, 1]$ & $0.02\%$ \\
    \bottomrule
  \end{tabular}
\end{table}

\begin{table}[t]
  \caption{NE difference results for \textsc{CVR\_Large} with varying numbers of selected experts}
  \label{table:varying_expert_selection:Web}
  \centering
  \begin{tabular}{ll|ll}
    \toprule
    Selection & NE diff
    & Selection & NE diff \\
    \midrule
    top-$1$ of $1$ expert
      & $0.00\%$
      & top-$1$ of $3$ experts
      & $-0.01\%$ \\
    top-$1$ of $2$ experts
      & $0.01\%$
      & top-$2$ of $3$ experts
      & $-0.04\%$ \\
    top-$2$ of $2$ experts
      & $0.03\%$
      & top-$3$ of $3$ experts
      & $-0.03\%$ \\
    \bottomrule
  \end{tabular}
\end{table}

\subsection{Effect of Expert Architecture}

We examine the influence of expert capacity by varying the expert network architectures. Experts are implemented as MLPs with one or two hidden layers with sizes from $\{256,512,1024\}$. We conducted experiments on the \textsc{CVR\_Large} dataset and use \textsc{Linear} layer as the routing module. Results are shown in Table~\ref{table:varying_expert_architectures:Web}, where the NE diff is calculated from the NE of experts with layer dimensions $[256,1]$.

We observe that $[512,1]$ achieves the lowest NE in all architecture designs. This may be due to the fact that the output from the shared architecture is of dimension $512$ for each training example, which is then fed to the main task as well as the auxiliary task of our \textsc{HME-PFD}. We also observe that (i) decreasing the expert network size from $[512,1]$ indeed worsens the model performance, which might be due to the insufficient model capacity of a smaller expert network; (ii) increasing the expert network size from $[512,1]$ does not improve the model performance, which might be due to underfitting of the larger expert networks. This set of experiments suggests a task-dependent design of the expert architecture.

\subsection{Effect of the Number of Selected Experts}

We investigate the effect of selecting multiple experts per input on the \textsc{CVR\_Large} dataset with $p=10$ privileged features. Specifically, we vary the total number of experts $e \in \{1,2,3\}$ and also vary the number of experts selected for each input (up to $e$, top-$k$ routing). Note that the case of $e=1$ with top-$1$ routing corresponds to standard \textsc{PFD}, where no actual routing happens, and all training examples are fed into the same expert network, which is part of the student model for the knowledge distillation. Results of the NE difference are presented in Table~\ref{table:varying_expert_selection:Web}.

We observe that the configuration with $e=3$ experts consistently outperforms the single-expert case ($e=1$, \textsc{PFD}). This highlights the core benefit of \textsc{HME-PFD}: combining representations from multiple specialized experts is more effective than relying on a single monolithic network, especially when dealing with complex and partially available privileged features. Interestingly, the model performs better with $e=3$ experts than $e=2$ experts, suggesting that with a large number of privileged features ($p=10$), two experts may not provide sufficient capacity for meaningful specialization. Selecting the top-$2$ experts yield the best result, outperforming both the more restrictive top-$1$ selection and the more inclusive top-$3$ selection. This indicates that while combining experts is beneficial, forcing every input to be processed by all experts may introduce noise and diminish the returns of the specialization. 

\section{Conclusion}
\label{sec:conclusion}

We proposed \emph{Heterogeneous Mixture-of-Experts Privileged Feature Distillation} (\textsc{HME-PFD}), a knowledge distillation framework designed for the realistic scenario where privileged features are heterogeneous and only partially observed during training. Unlike prior privileged feature distillation methods that restrict privileged features to the teacher model, \textsc{HME-PFD} additionally leverages privileged features within the student training framework through a mixture-of-experts architecture, where routing decisions explicitly account for heterogeneity in supervision quality.

We introduced both hard-coded and learnable routing strategies to guide expert selection based on privileged feature availability and values. Extensive experiments on large-scale post-click conversion prediction tasks demonstrate that \textsc{HME-PFD} consistently improves student performance, as measured by normalized entropy, even when privileged features are sparse or incomplete. These results highlight the effectiveness of exploiting privileged features not only for label distillation, but also for adaptive traffic routing in heterogeneous training environments. Our experiments also suggest that the model performance is dependent not only on the routing module design, but also on the number of experts, the expert architecture, the number of selected experts, etc.


\bibliographystyle{assets/plainnat}
\bibliography{references}

\clearpage
\beginappendix
\suppressfloats[t]
\section{Scenario-Conditional Analysis of Auxiliary Distillation}
\label{appendix:scenario-distillation}

Different availability patterns and values of privileged features may
correspond to distinct training scenarios. These scenarios need not differ
only in supervision quality: the relationship between regular features and
teacher targets may itself be scenario-dependent. We give a population-level
analysis of this effect for the auxiliary distillation objective. The analysis
fixes the shared representation and teacher, and does not assert improved
optimization convergence or a risk guarantee for the deployed main head.

\subsection{Setup}

Let $X$ denote regular features, $M$ a privileged-feature availability mask,
and $X^\star_{\mathrm{obs}}$ the observed privileged information. Let
$G\in\{1,\ldots,K\}$ be a measurable scenario indicator constructed from
$(X^\star_{\mathrm{obs}},M)$, with $\Pr(G=k)>0$ for each $k$. For example,
scenarios may distinguish full, partial, and absent privileged-feature
coverage, or merge several availability patterns into one group. Different
patterns do not necessarily imply different conditional target distributions;
this is precisely the distinction quantified below.

Fix a shared representation $Z=b_\theta(X)\in\mathbb R^d$ and a teacher target
$T=t(X,X^\star_{\mathrm{obs}},M)\in[0,1]$. All expectations are under the same
joint population law of $(Z,G,T)$. The single auxiliary head receives only
$Z$. In an idealized hard-routing model, the router observes $G$ and selects
an auxiliary expert $h_G$, while each expert itself receives only $Z$.
The scenario indicator and auxiliary experts are used only during training;
the deployed main prediction head still operates without privileged inputs.

Define the squared distillation risks
\begin{align}
L_{\mathrm{single}}(h)
&=\mathbb E\big[(T-h(Z))^2\big],
\label{eq:scenario-single-risk}\\
L_{\mathrm{routed}}(h_1,\ldots,h_K)
&=\mathbb E\big[(T-h_G(Z))^2\big].
\label{eq:scenario-routed-risk}
\end{align}
Let $\mathcal A$ be the set of all measurable functions from
$\mathbb R^d$ to $[0,1]$, and define
\[
L_{\mathrm{single}}^*=
\inf_{h\in\mathcal A}L_{\mathrm{single}}(h),
\qquad
L_{\mathrm{routed}}^*=
\inf_{(h_1,\ldots,h_K)\in\mathcal A^K}
L_{\mathrm{routed}}(h_1,\ldots,h_K).
\]
Write
\[
\mu(z)=\mathbb E[T\mid Z=z],
\qquad
\mu_k(z)=\mathbb E[T\mid Z=z,G=k],
\]
so that $\mu_G(Z)=\mathbb E[T\mid Z,G]$. Versions of conditional means outside
the corresponding support do not affect any of the risks.

\subsection{The Value of Scenario Conditioning}

\begin{theorem}[Oracle gain from scenario-conditional distillation]
\label{thm:scenario-oracle-gain}
Under the setup above,
\begin{equation}
\boxed{
L_{\mathrm{single}}^*-L_{\mathrm{routed}}^*
=
\mathbb E\big[(\mu_G(Z)-\mu(Z))^2\big]
\ge 0.
}
\label{eq:scenario-oracle-gain}
\end{equation}
The inequality is strict if and only if
$\Pr\big(\mu_G(Z)\ne\mu(Z)\big)>0$.
\end{theorem}

\begin{proof}
For any $h\in\mathcal A$, conditional expectation gives
\[
\mathbb E[(T-h(Z))^2]
=
\mathbb E[(T-\mu(Z))^2]
+
\mathbb E[(\mu(Z)-h(Z))^2],
\]
because $\mathbb E[T-\mu(Z)\mid Z]=0$. Since $T\in[0,1]$, its conditional
mean is also in $[0,1]$, and hence
\[
L_{\mathrm{single}}^*=\mathbb E[(T-\mu(Z))^2].
\]
The same argument conditional on $(Z,G)$ yields
\[
L_{\mathrm{routed}}^*=\mathbb E[(T-\mu_G(Z))^2],
\]
attained by choosing $h_k=\mu_k$ on the support of each scenario.
Finally, decompose
\[
T-\mu(Z)
=
(T-\mu_G(Z))+(\mu_G(Z)-\mu(Z)).
\]
The cross term has zero expectation because
$\mathbb E[T-\mu_G(Z)\mid Z,G]=0$. Therefore,
\[
\mathbb E[(T-\mu(Z))^2]
=
\mathbb E[(T-\mu_G(Z))^2]
+
\mathbb E[(\mu_G(Z)-\mu(Z))^2],
\]
which proves \eqref{eq:scenario-oracle-gain}. A nonnegative squared random
variable has positive expectation exactly when it is positive with positive
probability, proving the strictness criterion.
\end{proof}

\subsection{Interpretation and Scope}

Define the scenario heterogeneity term
\[
\Delta_\theta
:=\mathbb E[(\mu_G(Z)-\mu(Z))^2]
=\mathbb E\big[\operatorname{Var}(\mathbb E[T\mid Z,G]\mid Z)\big].
\]
It measures differences in teacher conditional means across scenarios after
accounting for the regular-feature representation. When $\Delta_\theta>0$,
a single head that does not observe the scenario must fit a pooled target,
whereas correctly routed experts can fit the scenario-specific targets.
Differences only in conditional noise variance, with identical conditional
means, do not produce a positive value of this term. Likewise, if $G$ is
already determined by $Z$, conditioning on $G$ adds no oracle information.

The result provides a modeling rationale for privileged-feature-aware
expert specialization, rather than an unconditional advantage from adding
experts. Its oracle gain comes from access to scenario information: a
sufficiently expressive single auxiliary head receiving both $(Z,G)$ can
attain the same oracle risk as the routed experts.

For finite model classes, let $\mathcal H_{\mathrm{single}}\subseteq\mathcal A$
and $\mathcal H_{\mathrm{routed}}\subseteq\mathcal A^K$ be the permitted
single-head and expert classes. Define their nonnegative approximation gaps
\begin{align*}
A_{\mathrm{single}}
&:=\inf_{h\in\mathcal H_{\mathrm{single}}}
L_{\mathrm{single}}(h)-L_{\mathrm{single}}^*,\\
A_{\mathrm{routed}}
&:=\inf_{\mathbf h\in\mathcal H_{\mathrm{routed}}}
L_{\mathrm{routed}}(\mathbf h)-L_{\mathrm{routed}}^*.
\end{align*}
Then the exact class-level comparison is
\[
\inf_{h\in\mathcal H_{\mathrm{single}}}L_{\mathrm{single}}(h)
-
\inf_{\mathbf h\in\mathcal H_{\mathrm{routed}}}L_{\mathrm{routed}}(\mathbf h)
=
\Delta_\theta+A_{\mathrm{single}}-A_{\mathrm{routed}}.
\]
Thus positive scenario heterogeneity alone does not guarantee an advantage
for arbitrary finite-capacity experts. Finite-sample estimation, imperfect
routing, and optimization may further offset the population modeling gain;
these effects are not bounded by the present theorem. Learned routing or a
jointly trained representation also requires additional analysis.

Finally, lower auxiliary distillation risk does not by itself imply lower
true-label risk of the deployed main prediction head. Any benefit to that
head must transfer through the shared representation and should be evaluated
separately using ground-truth labels.

\section{Extensions to Other Routing Paradigms}
\label{appendix:paradigms}

In the main text, we primarily consider routing each input to the top-$1$ or top-$k$ experts based on routing scores, with a small $k$. Our design and implementation follow the sparsely gated top-$k$ MoE framework introduced in~\citep{ICLR17_Shazeer}, where the experts with the highest $k$ scores are selected for each input. This type of sparsely gated MoE framework restricts each input to a limited number of experts, significantly reducing computational overhead while often retaining the expressive power of mixture-of-experts (MoEs). Beyond sparsely gated MoEs, a variety of alternative routing paradigms have been proposed, in which the assignment between inputs and experts differs from hard top-$k$ selection. In this appendix, we provide additional experimental results and discussions for two such paradigms, i.e., soft MoEs~\citep{ICLR24_Puigcerver} and expert choice routing~\citep{NeurIPS22_Zhou}.



\subsection{Soft MoEs}

\begin{table}[t]
  \caption{NE difference results for \textsc{CVR\_Large} with varying experts in soft MoEs (Lower NE indicates better performance)}
  \label{table:soft:varying_experts:Web}
  \centering
  \begin{tabular}{lccc}
    \toprule
    \multicolumn{1}{c}{}
    & $2$ experts
    & $3$ experts
    & $4$ experts \\
    \cmidrule(r){2-2}
    \cmidrule(r){3-3}
    \cmidrule(r){4-4}
    Routing module & NE diff & NE diff & NE diff \\
    \midrule
    \textsc{Plain}        & $0.00\%$ & N/A & N/A \\
    \textsc{Linear}        & $-0.02\%$ & $-0.07\%$ & $-0.07\%$ \\
    \textsc{MLP}          & $-0.03\%$ & $-0.03\%$ & $-0.04\%$ \\
    \textsc{ResNet}       & $0.00\%$  & $-0.03\%$ & $-0.04\%$ \\
    \textsc{Attention}   & $0.02\%$  & $-0.06\%$ & $-0.02\%$ \\
    \bottomrule
  \end{tabular}
\end{table}

\begin{table}
  \caption{NE difference results for \textsc{CVR\_Large} with varying experts in expert choice}
  \label{table:expert_choice:varying_experts:Web}
  \centering
  \begin{tabular}{lccc}
    \toprule
    \multicolumn{1}{c}{}
    & $2$ experts
    & $3$ experts
    & $4$ experts \\
    \cmidrule(r){2-2}
    \cmidrule(r){3-3}
    \cmidrule(r){4-4}
    Routing module & NE diff & NE diff & NE diff \\
    \midrule
    \textsc{Plain}        & $0.00\%$ & N/A & N/A \\
    \textsc{Linear}        & $1.99\%$ & $2.02\%$ & $1.83\%$ \\
    \textsc{MLP}          & $1.96\%$ & $1.93\%$ & $1.90\%$ \\
    \textsc{ResNet}       & $2.08\%$ & $1.88\%$ & $1.81\%$ \\
    \textsc{Attention}   & $1.98\%$ & $1.98\%$ & $1.97\%$ \\
    \bottomrule
  \end{tabular}
\end{table}

The sparsely gated MoEs~\citep{ICLR17_Shazeer} rely on hard assignment between inputs and experts. In contrast, soft MoEs introduce a soft mixing mechanism that allows each expert to process a weighted combination of all inputs or tokens~\citep{ICLR24_Puigcerver}. In the soft MoE framework of~\citep{ICLR24_Puigcerver}, \emph{dispatch weights} and \emph{combine weights} are defined over sequences of tokens, enabling flexible expert–token interactions.

In our ranking and recommendation setting, however, (i) the privileged features used for routing form a fixed-dimensional vector rather than a sequence of tokens, and (ii) the input to each expert is a single embedding vector rather than a token sequence. As a result, the dispatch and combine weight formulation in~\citep{ICLR24_Puigcerver} is not directly applicable. Instead, we adopt the core idea of soft MoEs by allowing each input to contribute to all experts with learned weights. Specifically, we use the routing module to generate expert scores based on privileged features and treat these scores as soft weights in the loss aggregation stage. This design is more closely related to the classical MoE formulation with non-sparse gating functions~\citep{NC91_Jacobs}, with the key distinction that routing decisions are driven by privileged features rather than the input embedding itself.

Experimental results on \textsc{CVR\_Large} are reported in Table~\ref{table:soft:varying_experts:Web}. We observe that soft MoEs outperform the \textsc{Plain} baseline in 10 of the 12 configurations. Compared with the top-$1$ routing results in Table~\ref{table:varying_experts:Web}, the performance gains of soft MoEs are slightly reduced for $e=2$ and for three of the four routers at $e=3$. But when the number of experts increases to $e=4$, soft MoEs continue to achieve meaningful performance improvements, particularly when using linear routing modules.



\subsection{Expert Choice}
Classic methods such as sparsely-gated top-$k$ MoE~\citep{ICLR17_Shazeer} and its variants perform routing by choosing experts from the perspective of input tokens or input data points. These methods might lead to issues of load imbalance and under-utilization of experts, especially when the number of experts are high. In~\citep{NeurIPS22_Zhou}, the expert choice (EC) is proposed, where each expert picks top-$k$ tokens,
attempting to address issues of load imbalance and under-utilization of experts directly. Since each expert picks and only picks top-$k$ tokens in each mini-batch, each expert is exposed to the exact same number of data points during training.

Here, we adopt the idea of EC and apply it to our MoE with routing based on privileged features. One key difference is that in the original EC, the experts are choosing the input tokens of top-$k$ scores in each mini-batch of training examples, where a token is part of one training example. For example, a token is typically one patch of images in computer vision tasks or one word embedding in a token sequence in natural language processing tasks.
In our recommendation setting, each input to the MoEs is the embedding of the raw input of dimension $d$, which does not embody a concept of a token. Instead, we choose the
top-$k$ embedding in a mini-batch with the highest score for each expert. We define the number of data points in a mini-batch as $n$ and the number of experts $e$. The capacity $c$ is the target average number of experts per input. Then, the $k$ can be defined as
\begin{align}
k = \max(1, \text{int}(\frac{n \times c}{e})),
\end{align}
where $\text{int}(\cdot)$ is a function to ensure that $k$ is an integer, and maximum with $1$ ensures that in the extreme case at least one input is chosen for each expert. It is easy to observe that by design, each expert is exposed to the exactly same amount of inputs, i.e., $k$ inputs per mini-batch. Another observation is that each input may be routed to a variety of experts since it can appear in the set of inputs with top-$k$ highest scores in more than one expert. We show the results in Table~\ref{table:expert_choice:varying_experts:Web}, where the capacity $c$ is set to $1$ and we vary the number of total experts. At this capacity, test NE is $1.81\%$ to $2.08\%$ higher than for \textsc{Plain} across all 12 expert-choice configurations.

\end{document}